\documentclass[letterpaper, 10 pt, conference]{ieeeconf}  

\usepackage{pdfpages}
\usepackage{upgreek}

\usepackage{amsmath}
\usepackage{url}

\IEEEoverridecommandlockouts                              

\title{\LARGE \bf
A Starter's Kit for Concentric Tube Robots
}

\author{Kalina Bonofiglio, Wenpeng Wang, Ethan R. Wilke, Adri Rajaraman, and Loris Fichera 
\thanks{The authors are with the Department of Robotics Engineering, 
Worcester Polytechnic Institute, Worcester, MA, 01609, USA. Corresponding author: {\tt\small wwang11@wpi.edu}}%
}

\begin{document}

\maketitle
\thispagestyle{empty}
\pagestyle{empty}

\begin{abstract}
Concentric Tube Robots (CTRs) have garnered significant interest within the
surgical robotics community because of their flexibility, dexterity, and
ease of miniaturization.
However, mastering the unique kinematics and design principles of CTRs can be challenging for newcomers to the field. 
In this paper, we present an educational kit aimed at lowering the
barriers to entry into concentric tube robot research.
Our goal is to provide accessible learning resources for CTRs,
bridging the knowledge gap between traditional robotic arms
and these specialized devices.
The proposed kit includes (1) An open-source design and assembly
instructions for an economical (cost of materials $\approx$ 700 USD)
modular CTR; (2) A set of self-study materials to learn the basics of 
CTR modeling and control, including automatically-graded assignments.
To evaluate the effectiveness of our educational kit, we 
conducted a human subjects study involving first-year graduate
students in engineering.
Over a four-week period, participants --- none of whom had any prior
knowledge of concentric tube robots --- successfully built their first
CTR using the provided materials, implemented the robot's kinematics
in MATLAB, and conducted a tip-tracking experiment with an optical
tracking device.
Our findings suggest that the proposed kit facilitates learning and
hands-on experience with CTRs, and furthermore, it has the potential to
help early-stage graduate students get rapidly started with CTR research.
By disseminating these resources, we hope to broaden
participation in concentric tube robot research to a wider a more
diverse group of researchers.
\end{abstract}

\section{INTRODUCTION}
\begin{figure}
    \centering
    \includegraphics[width=1\linewidth, trim=-1cm 0 -1cm -1cm,clip]{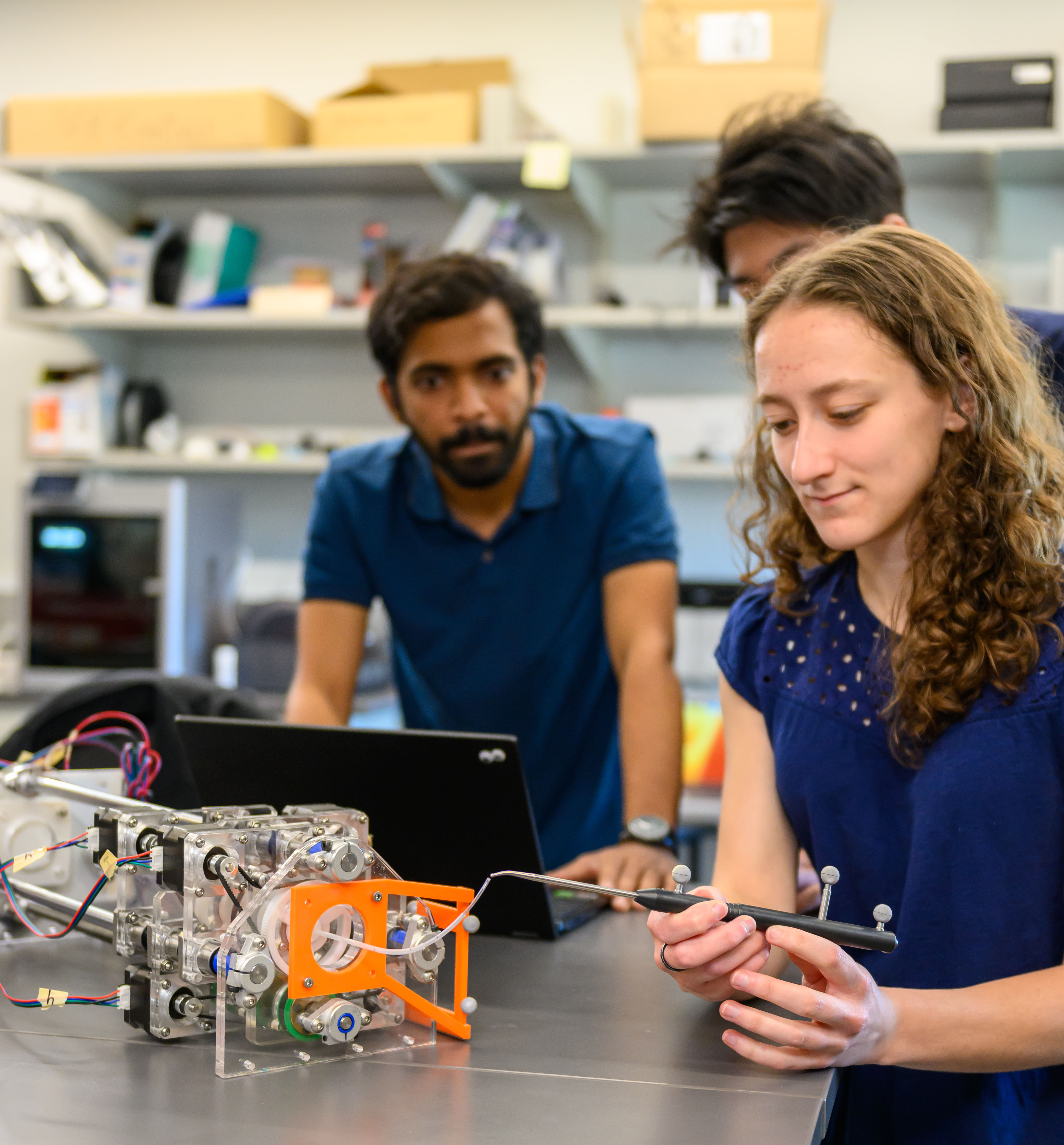}
    \caption{Authors of this manuscript measuring the position of 
    the tip of a concentric tube robot (CTR) 
    with an optically-tracked probe.
    The CTR pictured here is part of an educational
    kit developed by our team with the goal of
    providing accessible resources to learn
    CTR kinematics and acquire hands-on experience with these devices.
    }
    \label{fig:fig-1}
\end{figure}

Concentric tube robots (CTRs) are a class of thin,
flexible robotic arms designed to perform tasks in
constrained or hard-to-reach
environments~\cite{Nwafor2023,Mitros2022}.
The body of a CTR is composed of multiple pre-curved,
telescoping tubes. Each tube can be translated and
rotated independently of the others, producing
tentacle-like motions as
illustrated in Fig.~\ref{fig:fig-1}.
The minuscule form factor of these devices (prototypes 
as small as 430 $\upmu$m have been reported in 
the literature~\cite{Nwafor2023a}), combined with their
dexterity, makes them particularly well-suited for
use in medicine and surgery, with notable
applications including operations in the brain,
lungs, kidneys, and prostate.
A comprehensive survey of clinical applications of
CTRs to date can be found in the recent review article by 
Alfalahi \textit{et al}.~\cite{Alfalahi2020}.
\subsection{Barriers to Entry to Concentric Tube Robot Research}
Despite ever-increasing interest in CTRs,
significant barriers exist for researchers and
engineers who wish to contribute to the development
and clinical translation of these robots.
Learning how CTRs work is often the first barrier
to overcome.
Unlike \textit{conventional} robotic arms (i.e., arms 
made of rigid links interconnected by motorized joints),
for which a vast amount of self-study resources are
readily available in the form of textbooks, computer
simulators, and educational hardware platforms, 
few --- if any --- educational resources exist
for CTRs, which have not even made their way into
robotics textbooks yet.
In principle, aspiring learners could turn to the 
wealth of research manuscripts available in
the peer-reviewed literature for self-study.
Research manuscripts, however, are usually written with
a specialized audience in mind, potentially rendering
them intricate to comprehend for a broader readership. 
The use of technical language, combined with the dry, 
succinct descriptions of complex modeling/control
methodologies typical of research papers can be
a significant barrier to understanding for those
who are not already familiar with CTRs.
Furthermore, access to research manuscripts can
sometimes be constrained by paywalls, limiting their
circulation and inhibiting dissemination to a wider
and more diverse audience.
There is also a lack of accessible CTR hardware,
which restricts the ability of those new to the field
to gain crucial hands-on experience with these devices.
As of this writing, no commercially available
CTRs exist.
Open source designs are not available either,
with the only notable exception of the device of
Grassman \textit{et al}.~\cite{Grassmann2024}, who
recently proposed a mechanical platform to
construct various types of continuum robots,
including CTRs.
Most of the research groups active in CTR research
develop their own custom prototypes: In their
recent review of the literature, Nwafor and
colleagues identified 64 different CTR
prototypes~\cite{Nwafor2023}.
In a separate study, Grassman and
colleagues found that
many of these robots are used for a
single study or publication~\cite{Grassmann2020},
which suggests
that researchers --- probably due to the ever-present
pressure to publish --- frequently prefer to
develop quick prototypes, specifically tailored for a
given study, in place of modular, reusable
hardware that might be shared with the research
community.
A third barrier lies in the 
construction of the robot itself, which can be complex 
and require expensive materials and equipment.
The body of a CTR, in particular, is normally made of tubes
of Nickel-Titanium (Nitinol), a metal alloy known for its
ability to withstand large amounts of strain without undergoing plastic
deformation~\cite{Gilbert2016, dupont2009design, torabi2014compact}.
Such \textit{super-elastic} behavior is indeed a desirable
property for a CTR, as it enables the robot to have considerable
dexterity and cover large workspaces.
Despite that, Nitinol can be expensive to source (based on our
own experience, the typical cost of stock Nitinol tubing in the
United States is currently 50-100 USD per foot,
while the cost of custom manufacturing runs can be in the order
of tens of thousands of dollars).
Furthermore, Nitinol can be a challenging material to work with:
To enable tentacle-like motion, Nitinol tubes must first be formed
into prescribed curved shapes through an annealing treatment,
which, as previous work
has shown, can be time-consuming and
error-prone~\cite{Gilbert2016}.
Correct execution of this process requires specialized expertise
as well as equipment that may not be readily available in a
robotics laboratory.
\subsection{Contributions of this Study}
To eliminate these three barriers, and ultimately broaden
participation to CTR research, we propose a starter’s kit for concentric tube robots in this letter.

The proposed kit includes:
\begin{itemize}
    \item An open-source design and assembly instructions
    for an economical (cost of materials $\approx$ 700 USD), modular CTR.
    To keep costs low, our design capitalizes on the recent findings
    of Lu \textit{et al}.~\cite{Lu2023}, who demonstrated the viability of 
    CTRs made of Nylon instead of the more expensive Nitinol.
    \item A set of self-study materials to learn the basics of CTR modeling
    and control.
    These materials include self- and auto-graded assessments to help
    learners test their own understanding of the materials.
\end{itemize}
We report the results of a human subjects
study showing the effectiveness
of our proposed educational kit. Over a four-week period,
participants — graduate engineering students at our institution
with no prior knowledge of concentric tube robots — successfully built
their first CTR using the provided materials, implemented
the robot’s kinematics in MATLAB (The Mathworks, Natick, MA, USA),
and conducted a robot tracking experiment.

\section{Educational Kit Design}
The proposed educational kit can be
retrieved at:
\url{https://bit.ly/ open-source-ctr}.
We designed the kit around three main learning
objectives, i.e., to enable learners to
(1) build their first Concentric Tube Robot (CTR),
(2) mathematically describe the kinematics of CTRs,
and (3) analyze a CTR's accuracy and
identify sources of error.
Learners are expected to possess a basic level of 
proficiency in rapid prototyping and computer
programming.
Learners are furthermore expected to be familiar
with fundamental concepts  in robot kinematics,
including joint and task space,
mathematical representations of position and orientation,
and rigid-body motions.
We note that these topics are covered in virtually
every university-level robotics engineering curriculum,
including the one offered at our 
institution~\cite{Padir_2011}.
Additionally, a large body of educational
materials on these subjects is widely accessible
through various digital platforms and libraries.
\subsection{Learning the Kinematics of Concentric Tube Robots}
\begin{figure*}
\centering
\includegraphics[width=\textwidth]{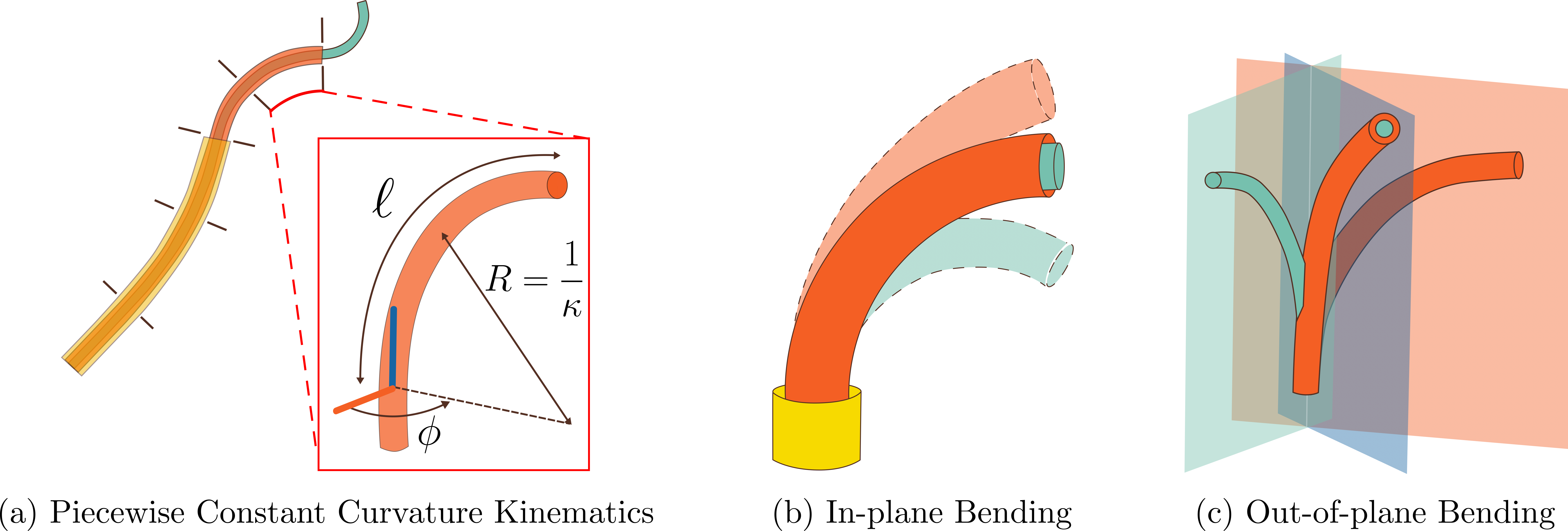}
    \caption{In our educational kit, we 
    refer to the kinematic modeling proposed in~\cite{webster_mechanics_2009}, where 
    (a) the shape of a 
    Concentric Tube Robot is partitioned into a 
    sequence of constant curvature arcs called \textit{links}.
    The geometry of each link is uniquely described by three
    parameters, namely the arc length $\ell$, the
    curvature $\kappa$ (or, alternatively, the radius of curvature $R$), and a rotation $\phi$ about the link's
    base.
    (b) 
    The curvature of each link is a weighted
    superposition of the natural curvatures of the 
    tubes present in the link, as captured by
    Eq.~\eqref{eq: in-plane-bending}.
    (c) If two (or more) concentric tubes do not share the same
    bending plane, the corresponding link will bend 
    \textit{in the middle}, as described by 
    Eqs.~\eqref{eq:out-of-plane-bending-1} and 
    \eqref{eq:out-of-plane-bending-2}.
    }
    \label{fig:self-study-materials}
\end{figure*}

To describe the kinematics of CTRs, our educational 
materials follow 
the approach developed by Webster
\textit{et al}.~\cite{webster_mechanics_2009}, which we
briefly review in the paragraphs below.
At its core, this approach uses Bernoulli-Euler beam theory
to describe the physical interactions occurring among the
tubes in a CTR and the shape that the robot will assume
as a result.
Although more advanced models have been
proposed in the literature (including more modern approaches based
on Cosserat Rod Theory~\cite{Mitros2022}), we opted to use
the kinematics from~\cite{webster_mechanics_2009}
as it is elegant, relatively straightforward to implement,
and therefore well-suited for learners with no prior CTR experience.
In addition, the derivation of CTR kinematics offered
in~\cite{webster_mechanics_2009}
introduces fundamental concepts that
are foundational for understanding more advanced models.
Our treatment of CTR kinematics is 
structured in three conceptual units, as illustrated in
Fig.~\ref{fig:self-study-materials}.
We begin by introducing the fundamental modeling assumption made
in~\cite{webster_mechanics_2009}, namely that the shape of a CTR
can be broken down into a sequence of constant curvature
arcs (called \textit{links}), each corresponding to unique
regions of overlap between tube sections.
Under this assumption, the geometry of each link
can be fully described by means of three parameters, namely, the link length 
$\ell$, the curvature $\kappa$, and a rotation $\phi$ about the base (see Fig.~\ref{fig:self-study-materials}(a)).
With these parameters, the homogeneous transformation matrix between
the base and the tip of a link can be expressed as
\begin{equation}
\mathbf{T} = 
\begin{bmatrix}
\mathbf{R} & \mathbf{p} \\
\mathbf{0}_{1\times3} & 1 
\end{bmatrix},
\end{equation}
where the rotation matrix $\mathbf{R}$ and the
translation vector $\mathbf{p}$ are given by, respectively,
\begin{equation}
\mathbf{R} = 
\begin{bmatrix}
\cos(\phi) \cos(\kappa \ell) & -\sin(\phi) & \cos(\phi) \sin(\kappa \ell)  \\
\sin(\phi) \cos(\kappa \ell) & \cos(\phi) & \sin(\phi) \sin(\kappa \ell)  \\
-\sin(\kappa \ell) & 0 & \cos(\kappa \ell) 
\end{bmatrix},
\end{equation}
\begin{equation}
\mathbf{p} = 
\begin{bmatrix}
\frac{\cos(\phi) (1-\cos(\kappa \ell))}{\kappa} \\
\frac{\sin(\phi) (1-\cos(\kappa \ell))}{\kappa} \\
\frac{\sin(\kappa \ell)}{\kappa} 
\end{bmatrix}.
\end{equation}
The number of links in a CTR and the corresponding link lengths
are determined by (1) the number and the geometry of the component tubes, and
(2) the translational position of each tube.
CTR are traditionally built using 
$n$ curved tubes, each having a straight
transmission section, resulting in $2n-1$
links (see Fig.~\ref{fig:self-study-materials}).
The curvature of each link is calculated as 
a weighted superposition of the tubes' curvatures at rest.
To illustrate this concept, consider the scenario
illustrated in Fig.~\ref{fig:self-study-materials}(b),
where two concentric tubes bend in the same plane.
From~\cite{webster_mechanics_2009}, the \textit{equilibrium}
curvature $\kappa$ is given by
\begin{equation}
\kappa = \frac{E_1 I_1\kappa_1 + E_2 I_2\kappa_2}{E_1 I_1 + E_2 I_2}
\end{equation}
where $E$ is the Young's modulus of the material, $I$ is the second moment
of area, and $\kappa$ is the tube's curvature at rest.
By extension, the curvature of a link with $n$ overlapping
tubes is
\begin{equation}
\kappa =\frac{\sum_{i=1}^n E_i I_i\kappa_i}{\sum_{i=1}^n E_i I_i}.
\label{eq: in-plane-bending}
\end{equation}
For the more general case where two or more concentric tubes do not
share the same bending plane, as illustrated in
Fig.~\ref{fig:self-study-materials}(c), 
the equilibrium bending plane can be found by calculating the 
corresponding rotation angle $\phi$:
\begin{equation}
\phi=\tan^{-1}\left(\frac{\chi}{\gamma}\right),
\label{eq:out-of-plane-bending-1}
\end{equation}
where $\chi$ and $\gamma$ are given by:
\begin{equation}
\chi=\frac{\Sigma_iE_iI_i\kappa_i\cos(\theta_i)}{\Sigma_iE_iI_i},
\gamma=\frac{\Sigma_iE_iI_i\kappa_i\sin(\theta_i)}{\Sigma_iE_iI_i}.
\label{eq:out-of-plane-bending-2}
\end{equation}
To enable learners to test their own understanding of CTR kinematics,
we provide a set of programming-based assessments.
These assessments use MATLAB Grader~\cite{MatlabGrader} as the
platform for submission and
evaluation. Learners' submissions are compared to a reference 
solution, which remains concealed, and constructive feedback is 
provided on erroneous entries, with the goal of facilitating
independent learning.

\subsection{Open-source Concentric Tube Robot Design}
\label{sec:design}
\begin{figure}
    \centering
    \includegraphics[width=1\linewidth]{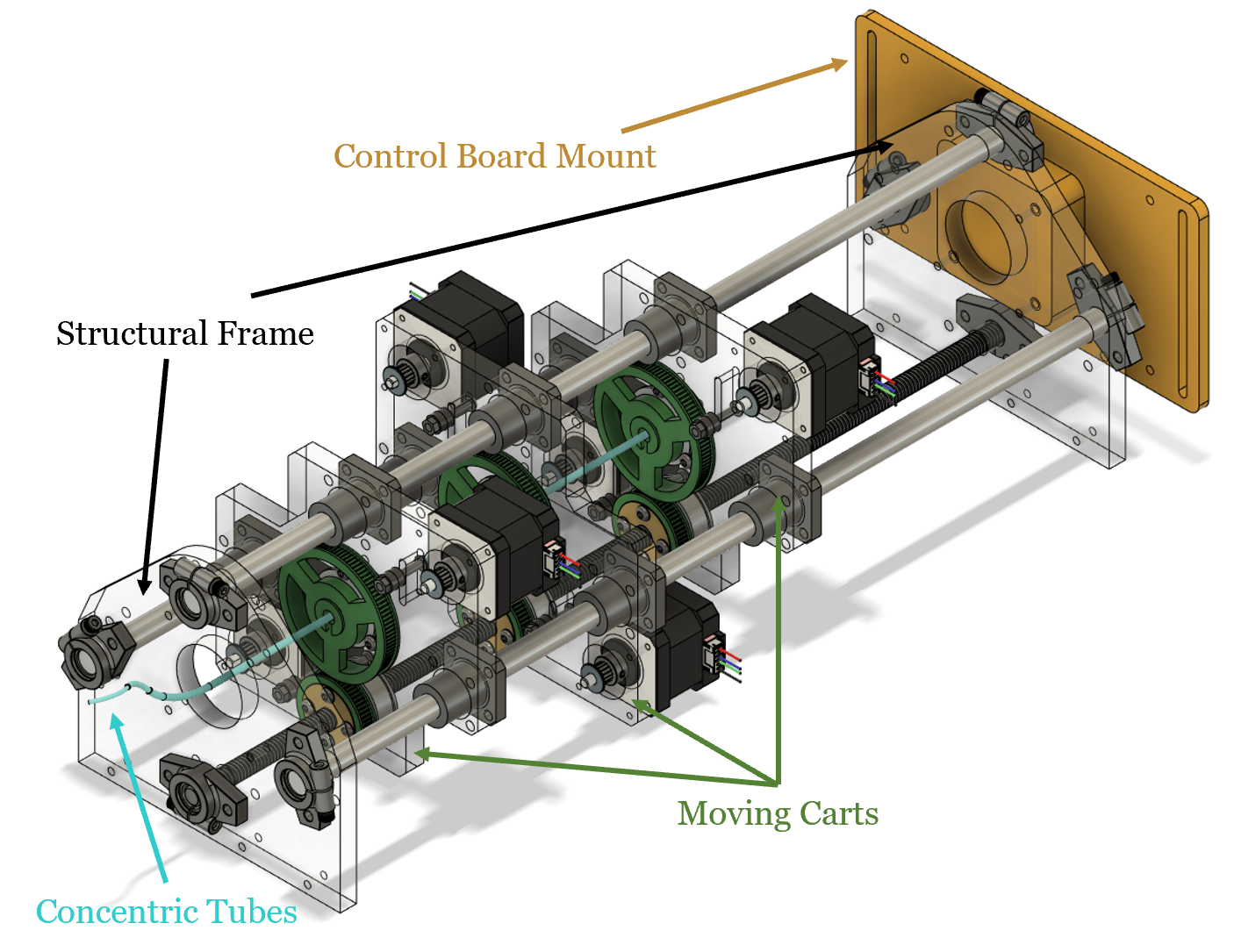}
    \caption{Proposed open-source concentric tube robot.
    Each tube is attached to a motorized
    cart moving on a system of rails.
    Each cart is equipped with two motors, respectively responsible
    for tube translation and rotation.
    Translation uses a rotating ball nut mechanism, with a fixed lead screw
    shared by all carts.
    }
    \label{fig:robot-system}
\end{figure}
\begin{figure*}
    \centering
    \includegraphics[width=\textwidth]{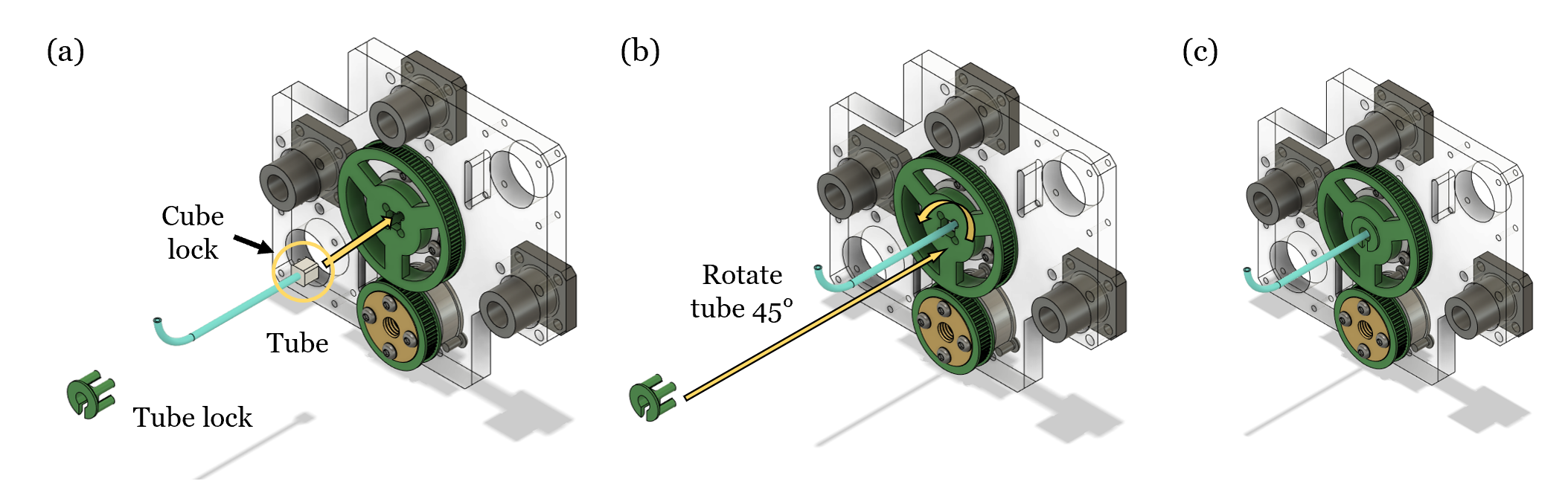}
    \caption{Procedure for installing a tube into a cart. (a) First, the cube 
    lock at the base of a tube is inserted into the green pulley. (b) The tube
    is rotated by 45$^\circ$ in either direction. A tube lock is inserted to hold the 
    cube in place. (c) The tube is now securely fastened. Tubes can be quickly
    removed or swapped by reversing this procedure.}
\label{fig:assembly-sample}
\end{figure*}
To gain hands-on experience with CTRs, 
learners can build the robot shown in Fig.~\ref{fig:robot-system},
for which we provide technical drawings,
detailed manufacturing and assembly instructions (an example of
which is shown in Fig.~\ref{fig:assembly-sample}), a complete Bill 
of Materials (BOM), and starting code.
As we shall see in the following sections (and, in greater detail,
in the BOM available on the website), the proposed design
uses mostly off-the-shelf components that can be readily purchased
from virtually any industrial supply store. Many of these components,
including the motors, the control board, and the lead screws/nuts
are the same that one would purchase to build a custom 3D printer.
There also is a limited number of custom components whose fabrication
requires access to a 3D printer.
\subsubsection{Actuation Unit}
\label{sec:actuation-unit}
\begin{figure}
    \centering
    \includegraphics[width=1\linewidth]{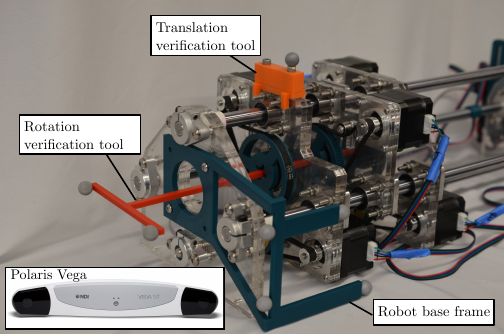}
    \caption{Experimental setup for the characterization of the motion accuracy of 
    the proposed actuation unit. An optical tracker, the Polaris Vega, is used to track
    the movement of the carts relative to a fixed frame of reference (the robot base frame).
    A rotation verification tool is used in place of a tube to track rotation. }
    \label{fig:system_verification}
\end{figure}

The actuation unit provides motion to the concentric tubes using a
system of carts moving on rails, as illustrated in Fig.~\ref{fig:robot-system}.
We note that actuation units similar to the one proposed here are prevalent in the CTR
literature~\cite{Nwafor2023}. Our design includes a tube quick-swap mechanism,
illustrated in Fig.~\ref{fig:assembly-sample}, whose purpose is to enable
the installation/removal of tubes from the robot without the need for
any tools.
Each cart is equipped with two NEMA 17 stepper motors 
(OMC Corporation Ltd, Nanjing City, PRC), which provide
tube translation and rotation.
Control of the stepper motors uses the Octopus V1.1 motion
control board (BigTreeTech, Shenzhen, PRC).
Position commands are sent to the motors in the form of
G-code strings, a well-known language for the control of
Computer Numerical Control (CNC) machines, 3D printers,
and other precise positioning devices.
We provide a MATLAB wrapper around this low-level G-code
interface, which allows users to rapidly write robot control
routines in a higher-level programming language.
To characterize the motion accuracy of our proposed
actuation unit, we carried out an experiment using the setup
illustrated in Fig.~\ref{fig:system_verification}.
Absolute position commands were sent to the control board,
and cart translation and rotation were captured with an
optical tracker, the Polaris Vega ST (NDI, Waterloo, ON, Canada).
We repeated this procedure 90 times, each time sending
target translation and rotation sampled at random in the
[0,50] mm and [-180\textdegree, 180\textdegree] range, respectively.
We observed a Root-Mean-Square Error (RMSE) of 0.10 mm for
translation and 0.08\textdegree~for rotation.
\subsubsection{Concentric Tubes}
\begin{figure*}
\label{fig: tubes}
    \centering
    \includegraphics[width=0.9\textwidth, trim=-1cm 0 -1cm -1cm,clip]{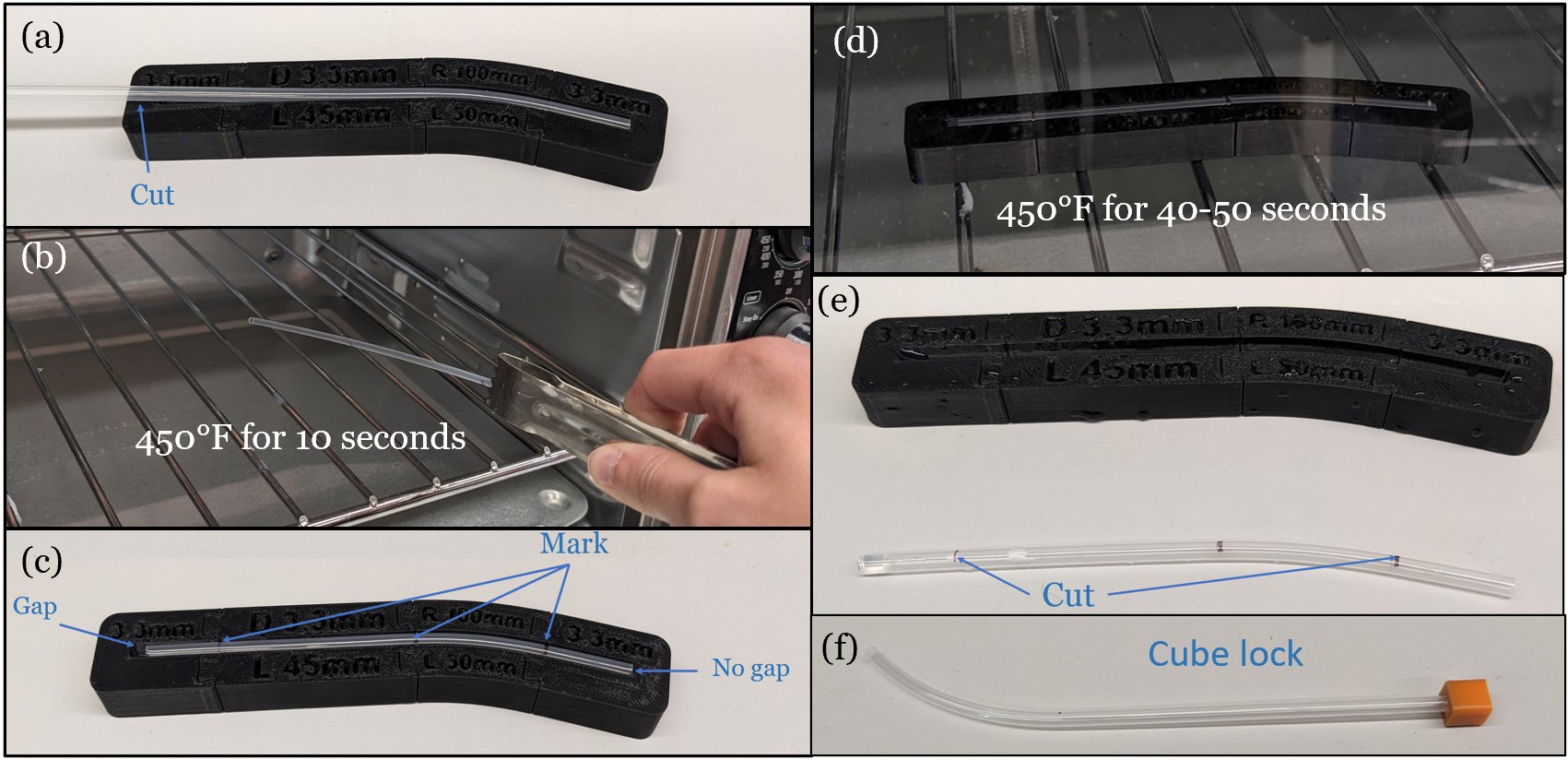}
    \caption{Procedure for shape setting the Nylon-12 tubes, adapted from~\cite{Lu2023}.
    (a) A Nylon-12 tube of the desired diameter is cut to length. (b) The tube is pre-heated for 10 seconds using a toaster oven set to 450$^\circ$F, for 10 seconds. (c) The tube is placed into the mold and transition points are marked with a felt tip pen. (d) The mold is now placed into the toaster oven for 40-50 seconds, then removed and quenched into a cool water bath. (e) The tube is removed from the mold and cut where indicated, using a sharp razor to avoid creasing. (f) A cube lock is glued onto the straight end of the tube. The tube is now
    ready to be installed in the robot.}
    \label{fig:shape-setting}
\end{figure*}

Our educational CTR uses 
tubes made of Nylon-12, a grade of Nylon commonly
employed in medical devices.
To build the prototype shown in Fig.~\ref{fig:fig-1},
we sourced tubing from Duke Extrusion
(Morgan Hill, CA, USA).
For the shape setting of the tubes, we adopt the protocol recently proposed
by Lu \textit{et al.}~\cite{Lu2023}, which uses heat to soften the tubes and
molds to impose prescribed curvatures. For completeness, we illustrate this
protocol in Fig.~\ref{fig:shape-setting}.
As part of the kit posted on our website, we provide the CAD files necessary to create
the molds, which can be either 3D printed or laser cut.
\subsubsection{Cost Analysis}
Table~\ref{tab:system_costs} reports a cost breakdown
for the materials necessary to build the proposed robot,
based on the BOM posted on our website.
Listed amounts are in 2024 US dollars. Data is based
on the advertised retail price of individual items
(current as of the submission date of this manuscript), 
excluding any applicable sales tax. 
Aggregate amounts for each category are rounded up to 
the nearest tens of dollars.
\begin{table}
\centering
\caption{Cost Breakdown for the Proposed Open-Source Concentric Tube Robot}
\label{tab:system_costs}
\begin{tabular}{lc}
\hline
\textbf{Component}        & \textbf{Cost}      \\
\hline
Mechanical Components (Lead Screws, Rods, Bearings, Belts) & \$250 \\
Hardware (Screws, Nuts, Bolts, Washers)              & \$90  \\
Electronics \& Motors & \$180 \\
Raw Materials (Tubes, Acrylic)         & \$180 \\
\hline
\textbf{Total}            & \$700  \\
\hline
\end{tabular}
\end{table}

\section{Human Subjects Study}
We conducted a human subjects study where participants 
built their first CTR and modeled its kinematics with the aid
of our educational kit.
The study was approved by the Institutional Review
Board (IRB) of our university, under protocol number 24-0020.
We recruited a total of six participants among the population of 
graduate students at our institution. Specific inclusion criteria
were (1) previous degree in either engineering or science, (2) 
foundational knowledge of robot kinematics, and (3) no prior 
knowledge or experience
with CTRs.
\subsection{Methodology}
\begin{figure*}
    \centering
    \includegraphics[width=\textwidth]{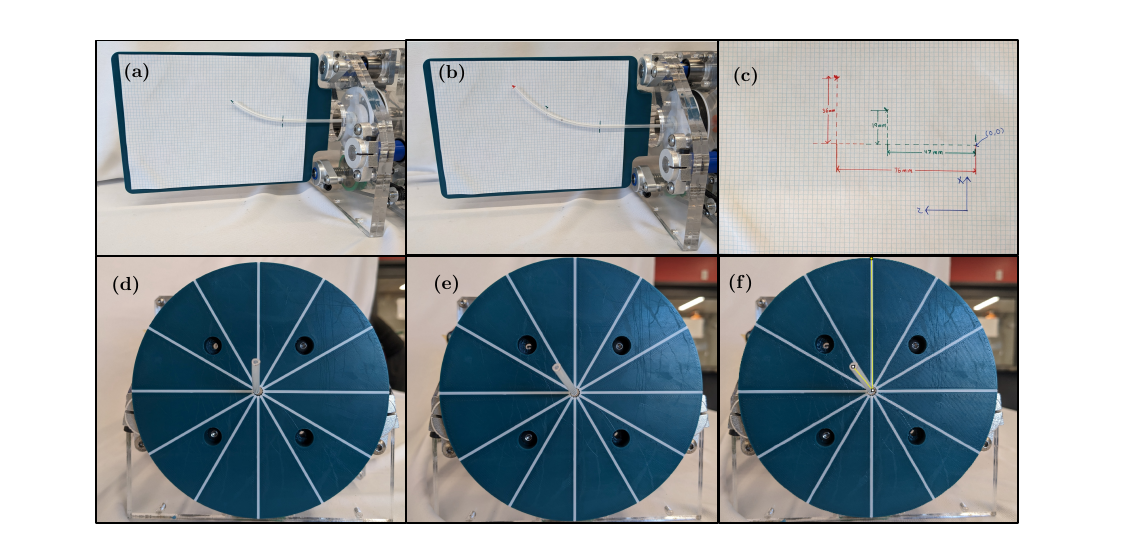}
    \caption{(Top row) In-plane translation experiment: (a) Initial configuration; the tip
    position is marked on engineering paper with a green marker.
    (b) The inner tube is translated forward and the new tip position is marked in red. 
    (c) CTR tip position coordinates before (green) and after (red) the translation of the inner tube.
    Participants in the study were asked to verify agreement between these coordinates
    and the output of the kinematics model they had implemented in week 3.
    (Bottom row) Out-of-plane rotation experiment: 
    (d) Initial configuration. (e) The outer tube is rotated. 
    (f) The new orientation is measured by digitizing points in the image, using
    ImageJ~\cite{Rueden2017}. Analogously to the previous experiment, participants 
    were asked to compare the new orientation with the output of their kinematics model
    and verify agreement.
    }
    \label{fig:userStudy_experiments}
\end{figure*}

\begin{figure*}
    \centering
    \includegraphics[width=\textwidth]{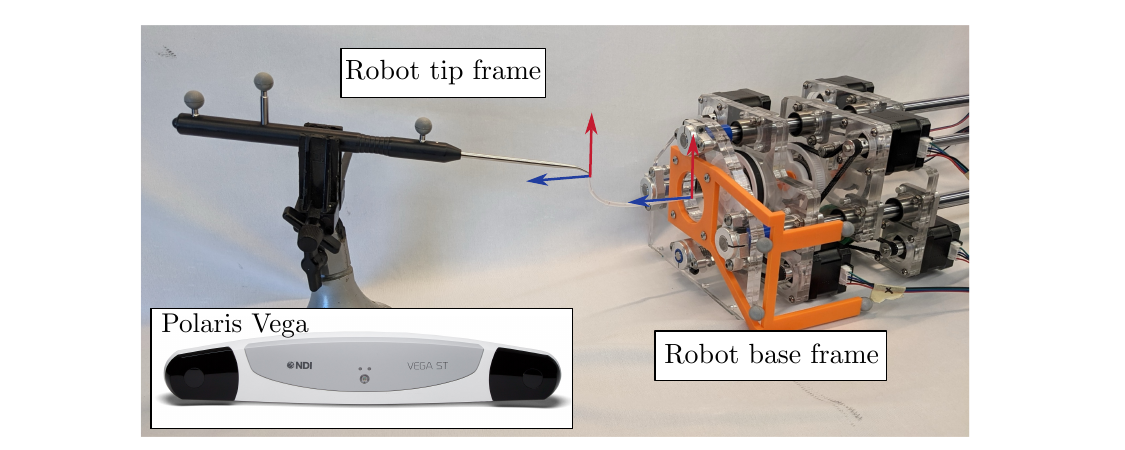}
    \caption{Kinematic verification workspace with frames of interest obtained by Polaris Vega. (Robot tip frame): the tip of the concentric tubes or end effector frame obtained by marker probe. (Robot base frame): defined at the center of structural frame cut-out, which can be obtained through CAD parameters of the orange 3D printed component.}
    \label{fig:fig-transform}
\end{figure*}

The study lasted a total of four weeks, as detailed in Table
\ref{tab:timeline}. 
\begin{table}
\centering
\caption{Human Subjects Study Timeline}

\begin{tabular}{| l |  l |} 
 \hline
\multicolumn{2}{|c|}{\textbf{Week 1}} \\
 \hline\hline
\textbf{Preparatory Work} & $\circ$ Laboratory safety training\\
  &   $\circ$ Pre-study survey \\ 
 \hline
 \textbf{Actuation Unit Build} &  $\circ$ Manufacturing of custom parts\\ 
  & (laser cutting \& 3D printing) \\
  &   $\circ$ Actuation Unit Assembly \\
 \hline
 \hline
\multicolumn{2}{|c|}{\textbf{Week 2}} \\
 \hline
 \hline
 \textbf{Tube Manufacturing} &  $\circ$ Manufacturing of the tube molds\\ 
  &    $\circ$ Shape setting of the tubes \\
 \hline
 \textbf{Control System Setup}  &  $\circ$ Configuration of the NEMA 17 board\\ 
  &   $\circ$ Actuation unit testing \\
 \hline
 \hline
\multicolumn{2}{|c|}{\textbf{Week 3}} \\
 \hline
 \hline
 \textbf{CTR Kinematics} &  $\circ$ Studying the provided reading materials\\ 
  &   $\circ$ Complete MATLAB grader assignments \\
 \hline
 \hline
\multicolumn{2}{|c|}{\textbf{Week 4}} \\
 \hline
 \hline
 \textbf{Experiments} &  $\circ$ In-plane bending experiment\\ 
  &   $\circ$ Out-of-plane bending experiment \\
  &   $\circ$ Experiment with optical tracker \\
 \hline
\textbf{Post Study Work}  &  $\circ$ Post-study survey\\
 \hline
 
\end{tabular}
\label{tab:timeline}
\end{table}
During the first two weeks, participants were
tasked with building a CTR based on the
instructions provided in our educational kit.
This activity was carried out in teams, with participants being
randomly assigned to one of two groups.
Once robot assembly was complete, participants performed a
motion accuracy test, identical to the one described earlier
in section~\ref{sec:actuation-unit}, to ensure that their robot's 
actuation unit responded to commands repeatably and accurately. 
In the third week, participants studied the kinematics of CTRs and
developed a library of MATLAB functions with the aid of our
MATLAB Grader assessments.
The final week of the study was dedicated to robot tracking
experiments, whose purpose was to give participants the opportunity 
to observe actual CTR kinematics at play.
The first two experiments are illustrated in Fig.~\ref{fig:userStudy_experiments}:
In these experiments, participants operated a simple CTR with only two tubes and 
observed the mechanics of in-plane and out-of-plane bending, comparing observations
with the predictions of the kinematic model they had previously implemented in MATLAB.
In the third and final experiment, illustrated in Fig.~\ref{fig:fig-transform}, participants
performed tracking of 3-tube CTR using
an optical tracker. 
Throughout the study, we documented participant learning through a combination
of objective and subjective forms of evidence.
Overall progress was monitored through weekly check-ins
wherein we verified whether participants had been
able to attain all the milestones for the week.
Learning of CTR kinematics in Week 3 was documented through MATLAB
Grader, which provides instant metrics on how many learners
have completed each assignment at any given time and how 
long it took to solve a specific problem set.
In addition to the above, we administered questionnaires where
participants were asked to rate the their confidence in their ability to (\textbf{Q1}) build a CTR, (\textbf{Q2}) describe its principle of operation, (\textbf{Q3}) control the robot, and (\textbf{Q4}) calculate its forward kinematics. Also,
participants were asked to evaluate the clarity of the 
provided (\textbf{Q5}) construction and assembly guide for
the robot and (\textbf{Q6}) fabrication guide for the concentric tubes.
Ratings used a 5-point Likert scale, with 1 being
"Not Confident/Not Clear" and 5 being "Very Confident/Very Clear."
\subsection{Results}
\begin{figure}
    \centering
    \includegraphics[width=\linewidth]{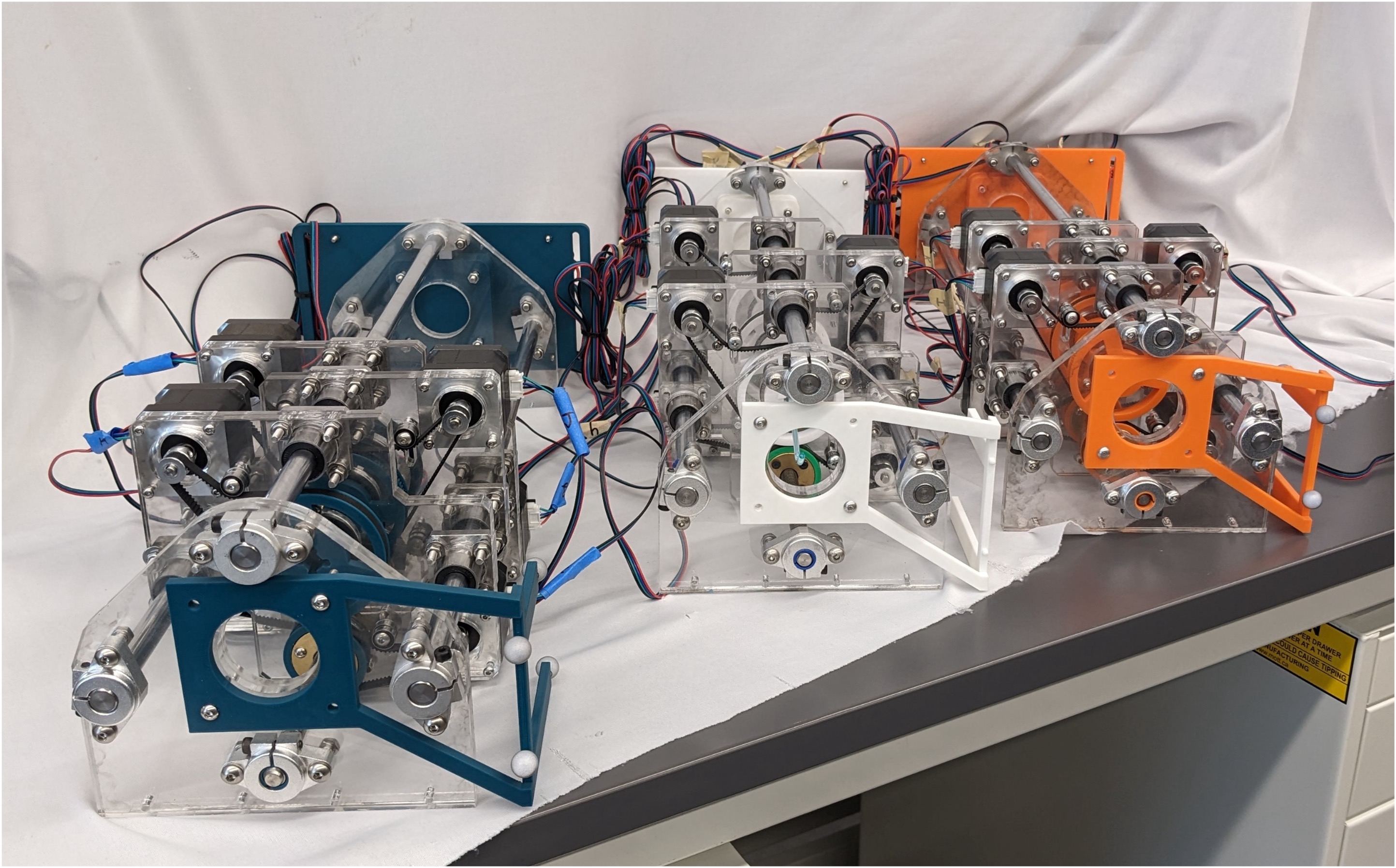}
    \caption{Actuation units built by participants in
    the first two week of the study.}
    \label{fig:3ctrs}
\end{figure}

\begin{figure*}
    \centering
    \includegraphics[width=0.85\textwidth, trim=-1cm 0 -1cm -1cm,clip]{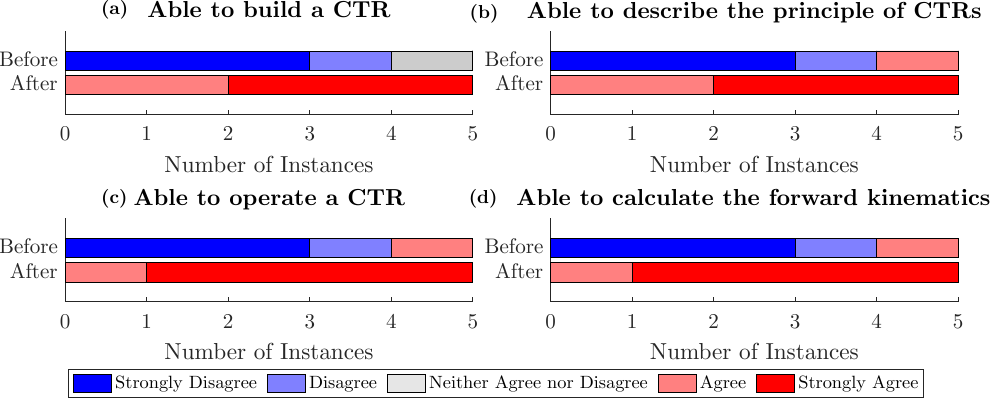}
    \caption{User reported CTR abilities before and after the study.}
    \label{fig:User-abilities}
\end{figure*}

All participants successfully completed the study,
with the exception of one participant who elected to 
leave during the third week, citing lack of time
as the main reason for their withdrawal.
Fig.~\ref{fig:3ctrs} shows the 
two robots built by study participants,
next to the prototype built earlier
by our team.
Table~\ref{tab:matlab-grader} summarizes the participants'
activity on MATLAB Grader during Week 3, when the main
task was to learn and implement CTR kinematics in MATLAB.
The metrics we report include the number
of participants who were able to correctly implement CTR
kinematics, and the average and maximum observed number of attempts for each programming assignment.
The first two assignments, namely the implementation of
models to capture in-plane and out-of-plane bending (refer 
to Fig.~\ref{fig:self-study-materials}), were relatively
straightforward to implement and only required an average of
two attempts on the part of participants.
The latter two assignments were moderately more complex 
and required more time to solve.
The third assignment, in particular, required participants
to implement an algorithm to calculate link lengths 
based on the translational position of the tubes.
The average number of attempts to solve this problem 
correctly was 4, with one participant making a total
of 13 attempts, and one participant --- the one who ultimately left the study --- unable to provide a correct
solution.
\begin{table*}[]
\centering
\caption{MATLAB Grader Results}
\begin{tabular}{rccc}
\textbf{Assignment Description} & \textbf{\begin{tabular}[c]{@{}c@{}}Number of Participants Who \\ Solved Assignment Correctly\end{tabular}} & \textbf{\begin{tabular}[c]{@{}c@{}}Average Number of Submissions \\ per Participant\end{tabular}} & \textbf{\begin{tabular}[c]{@{}c@{}}Maximum Number of Submissions \\ per Participant\end{tabular}} \\ \hline
Modelling In-Plane Bending                & 6                                                                                                          & 2                                                                                & 4                                                                                \\
Modelling Out-of-Plane Bending            & 6                                                                                                          & 2                                                                                & 3                                                                                \\
Calculating Link Lengths & 5                                                                                                          & 4                                                                                & 13                                                                               \\
Calculating the Full Kinematics              & 5                                                                                                          & 4                                                                                & 10                                                                              
\end{tabular}
\label{tab:matlab-grader}
\end{table*}
Fig.~\ref{fig:User-abilities} reports the responses
to questions \textbf{Q1}---\textbf{Q4} provided by the
five participants who completed the study.
We observed a significant increase in the participants'
confidence in their knowledge of CTR topics and
abilities related to CTRs.
All participants either agreed or strongly agreed that the assembly guides provided sufficient instruction to construct the device and that the kinematics assignments advanced their understanding of CTR kinematics 
(\textbf{Q4}---\textbf{Q5}).

\section{Discussion}
In this paper, we described an educational kit
for concentric tube robots consisting of an open-source
robot design and materials to study CTR kinematics.
The proposed robot design costs $\approx$ \$700,
making it the most cost-accessible, open-source
CTR design available to date.
We note that Grassman \textit{et al.}~\cite{Grassmann2024} recently proposed 
an open-source actuation unit that can be used to build CTRs.
While their device supports the implementation of more advanced control
algorithms, it also has a significant higher cost (1785 USD) than
the one proposed in this paper.
Results of our human subjects study appear to support the
claim that our educational kit has the potential to
help engineers learn the basics of CTR modeling and
build their first CTR system.
It is important to acknowledge the study involved a
small group size (n=6), recruited at a single
institution. While results from this group provide preliminary
indication that our proposed kit has the potential to expand
access to CTR research to a larger and more diverse pool
of researchers, we believe that a larger, multi-center study
would be required to better corroborate this claim.
Since out proposed CTR design is open-source, it can be easily
customized depending on a user's expertise and available budget.
For instance, a more advanced user could easily replace the Nylon-12
tubes with Nitinol tubes without having to make any changes to the 
hardware.
Additionally, stepper motors are an excellent and budget-friendly
choice of motor, but users that can afford to spend more on
motors could adapt the design by changing the motor
mounting holes on the cart plates to match their preferred motors.
By creating a modular device that starts a low price point,
our intent is to make this design accessible to the largest
possible number of users without preventing more established
or better-funded users from using this design as a
foundation for a more advanced system. 

\section{CONCLUSION}
In this work, we have presented an open-source CTR platform
design that includes sufficient documentation for replication
and a set of self-guided materials to introduce a newcomer
to CTR mechanics. The cost of our device is $\approx$ \$700,
making it the least expensive open-source CTR by a significant margin. 
Our extensive documentation allows others to exactly replicate our device, as we demonstrated with our human subjects study. We also developed learning materials that were sufficient to guide users through the implementation of a full CTR kinematics model and a set of experiments that demonstrate the operating principle of CTRs. While we did not yet conduct a full classroom experiment, the results of our user study show that this platform is a promising option for teaching students about the mechanics of CTRs. 
Using our foundational observations presented in this study, we envision expanding the existing user study into a multi-center study involving participants from a broader range of engineering disciplines and employing more objective methods for data collection. The anticipated results will enable a more accurate evaluation of the proposed platform and learning materials in terms of their effectiveness in teaching fundamental mechanics concepts to newcomers in the field of CTRs.

\bibliographystyle{IEEEtran}
\bibliography{References}

\end{document}